*Article*

# A Bio-Inspired Chaos Sensor Model Based on the Perceptron Neural Network: Machine Learning Concept and Application for Computational Neuro-Science

Andrei Velichko *, Petr Boriskov, Maksim Belyaev and Vadim Putrolaynen

Institute of Physics and Technology, Petrozavodsk State University, 33 Lenin str., 185910 Petrozavodsk, Russia; boriskov@petrsu.ru (P.B.); biomax89@yandex.ru (M.B.); vputr@petrsu.ru (V.P.)

* Correspondence: velichko@petrsu.ru

**Abstract:** The study presents a bio-inspired chaos sensor model based on the perceptron neural network for the estimation of entropy of spike train in neurodynamic systems. After training, the sensor on perceptron, having 50 neurons in the hidden layer and 1 neuron at the output, approximates the fuzzy entropy of a short time series with high accuracy, with a determination coefficient of $R^2 \sim 0.9$. The Hindmarsh–Rose spike model was used to generate time series of spike intervals, and datasets for training and testing the perceptron. The selection of the hyperparameters of the perceptron model and the estimation of the sensor accuracy were performed using the K-block cross-validation method. Even for a hidden layer with one neuron, the model approximates the fuzzy entropy with good results and the metric $R^2 \sim 0.5 \div 0.8$. In a simplified model with one neuron and equal weights in the first layer, the principle of approximation is based on the linear transformation of the average value of the time series into the entropy value. An example of using the chaos sensor on spike train of action potential recordings from the L5 dorsal rootlet of rat is provided. The bio-inspired chaos sensor model based on an ensemble of neurons is able to dynamically track the chaotic behavior of a spike signal and transmit this information to other parts of the neurodynamic model for further processing. The study will be useful for specialists in the field of computational neuroscience, and also to create humanoid and animal robots, and bio-robots with limited resources.

## 1. Introduction

Numerous studies in the field of neurophysiology and biophysics over the past 40 years showed that chaotic dynamics is an inherent property of the brain. Neuroscientists are interested in the role of chaos in cognition and the use of chaos in artificial intelligence applications based on neuromorphic concepts. The chaos is important for increasing the sensory sensitivity of cognitive processes [1–7]. For example, the sensory neuronal systems of crayfish and paddlefish use background noise to detect subtle movements in the water made by predators and victims [1]. Another study of the central nervous system [2] demonstrated that the efficient search for food using memory regions of the olfactory bulb of rabbits is also reflected by the chaotic dynamics of electroencephalography (EEG). Van der Groen et al. [4] investigated the effect of stochastic resonance on cognitive activity in the brain. The random stimulation of the visual cortex with an optimal noise level can be used as a non-invasive method to improve the accuracy of perceptual decisions. Diagnosis of chaotic and regular states, especially in physiology and medicine, can be of great practical importance. For example, some medical studies consider normal dynamics of health as ordered and regular, while many pathologies are deemed as bifurcations into chaos [8–10]. In that regard, the ventricular fibrillation, an arrhythmia most often associated with sudden death, is a turbulent process (cardiac chaos), which may be the result of the mechanism of subharmonic bifurcation (period doubling) [9]. On the other hand, a mild form of ECG chaos may indicate congestive heart failure [11]. Studying the orderliness of the ensemble functioning of brain neurons by measuring electroencephalograms makes it possible to: identify pathological conditions (neurodegenerative diseases, epilepsy) [12–14]; determine the level of concentration and mental fatigue [15–17]; evaluate the emotional state [18,19]; and produce biometric identification of a person [20,21].

Although scientists are interested in detecting chaos in biological systems to identify pathological conditions, and its influence on the dynamics of bioprocesses, a question may arise: *Can a simulated model of a biosystem evaluate the randomness of a spike train?* A possible answer to this question could be the presentation of a bio-inspired model of the chaos sensor that originates as the goal for this study. The paper proposes a bio-inspired tool for assessing chaos based on a multilayer perceptron (Figure 1). The question of whether an ensemble of neurons with a similar function actually exists in real life may be a topic for future research by neurophysiologists.

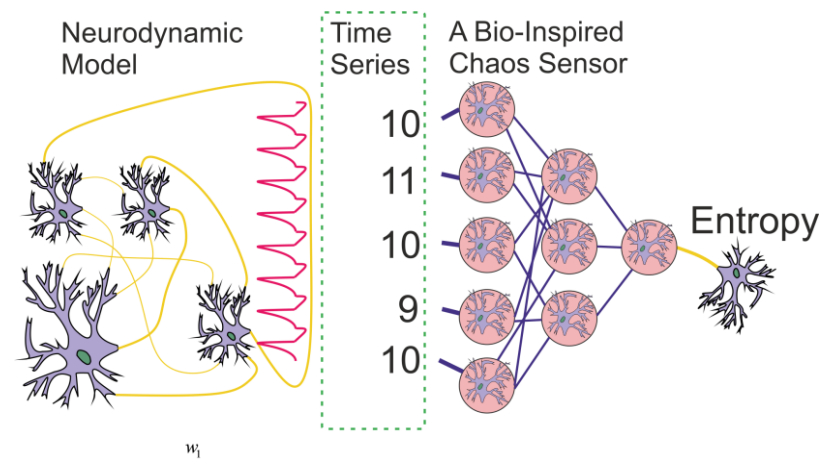

**Figure 1.** The concept of a chaos sensor based on a perceptron, where a spike signal is input as a time series of periods, and the signal entropy is estimated at the output neuron.

The central (cortical) nervous system (CNS) of the brain is highly heterogeneous. In fact, the neurons are weakly connected to each other, and each of them is directly affected by less than 3% of the surrounding neurons [22]. Moreover, neurons are individual and highly recurrent, that is, they have feedback through the neurons of the environment [23]. The CNS weakly resembles an artificial neural network (ANN) [24], where neuron nodes are functionally equal and each neuron receives a signal from all neurons of the previous layer.

In 1986, Rumelhart, McClelland and their research group introduced the term connectionism [25], which meant a set of algorithms and models with parallel-distributed data processing. The concept of connectionism imitates the basic computational properties of groups of real neurons found in the central nervous system using small network modeling of abstract units [26–28]. However, the first classical example of such small systems, proposed by Rosenblatt [29] long before the work of Rumelhart and McClelland, is the perceptron, the simplest neural network of direct propagation consisting of separate layers with one output neuron, which linearly separates (classifies) information into two categories. The Rosenblatt's perceptron, consisting of only one (hidden) layer, played an outstanding role in the development of neural technologies. The first ANN developments and the main learning method, the error correction method, known as the backpropagation method, were based on Rosenblatt's perceptron [30]. Perceptrons are the building blocks of more complex ANNs [24].

Multilayer perceptrons (MLP) can solve problems where their ability to quickly adapt to sudden changes in the environment is used. As an example, the ability of a robot based on a multilayer perceptron to move over rough terrain is analyzed in [31]. The study [32] indicates that multilayer perceptrons are good candidates for modeling the individual and collective behavior of fish, in particular, in the context of social biohybrid systems consisting of animals and robots. MLP collective behavior models are an interesting approach to design animal and robot social interactions. In [33], a biorobot with a Brain Machine Interface based on a MLP activated by rat brain signals, was developed. The results showed that even with a limited number of implanted electrodes connected to the neurons of the input layer of the perceptron, the signals of the rat brain are synchronized with high accuracy with the movement of the bio-robot.

In this paper, we propose the machine learning (ML) concept of a chaos sensor based on a perceptron. The concept of the sensor is presented in Figure 1. The neurodynamic model generates a spike signal, and the sensor evaluates its irregularity as an entropy value. The chaos sensor model evaluates the regularity of the spike oscillogram from a time series of periods between spikes. The time series is input to the perceptron, and the entropy value is calculated at the output. Such sensor can be applied not only in the field of computational neuroscience, but also to create humanoid robots [34] and animal robots [35,36], including devices with limited resources [37], as well as to improve neuroadaptive learning algorithms for robust control in constrained nonlinear systems [38,39]. For example, in biosimilar models where information is distributed as a spike train, changes in the packet frequency are often studied [40,41]. However, the entropy of the spike packet carries valuable information, and biosimilar chaos sensors can be used in signal processing both in robots and bioimplants. In this study, we show that the modes of skin stimulation and rest differ in the value of the entropy of the spike packets of action potential recordings from the L5 dorsal rootlet of rat. Therefore, the fact of stimulation can be detected by a perceptron-based biosimilar chaos sensor. Such sensor does not require much computing power, as the perceptron model is simple, takes up little RAM, and can be implemented on devices such as the Arduino Uno [42,43].

ML methods can fundamentally change the hardware designs of traditional sensors and can be used to holistically design intelligent sensor systems [44] and the Artificial Intelligence sensoric [45,46]. The ML sensor paradigm was further developed by Warden et al. [47] and Matthew Stewart [48], where the authors introduced the terms Sensors 1.0 and Sensors 2.0 devices. Sensors 2.0 devices are the combination of both a physical sensor and a ML module in one package. The MLP-based chaos sensor presented in this study belongs to the type of ML sensors, as it is trained by ML methods. If the chaos sensor would be implemented on the real device, it would be attributed to Sensor 2.0 devices. The input of the sensor will be a time series accumulated from the spike train, and the output will be the entropy value (Figure 1). In addition, the current development can be attributed to the bio-inspired ML sensors, as it is implemented on MLP and trained on neurodynamic models or spike datasets of real biological systems. Bioinspired sensor systems, in comparison to traditional sensing systems, use existing technologies and processes to simulate natural structures, materials and natural sensing models [49–51]. In addition to its high-sensitivity performance, rapid response, durability, and low-power consumption, it provides the sensing system with some specific properties, such as self-healing, self-cleaning or adaptability. As a result, bioinspired sensor systems have emerged as a hot research direction, including tactile sensors for electronic skin [52,53], chemical sensors for the bioelectronic nose [54] and bioelectronic tongue [55], auditory sensors for hearing aids [56] and other bioinspired electronics for artificial sensory systems based on the five traditionally recognized senses of sight, hearing, smell, taste, and touch [50,57]. We extend this list by presenting the chaos sensor for spike train in neurodynamic systems.

For the chaos studies, a reliable method for chaos detection is the key. The main characteristic of chaos is entropy: an additive physical quantity as a measure of the possible states of a system, introduced into the methodology of statistical physics and thermodynamics since the middle of the 19th century. In the 20th century, due to the significant use of the entropy concept in other areas of the natural sciences, including biology, mathematicians proposed numerous variations of this characteristic. A number of entropy indicators and calculation algorithms have appeared for objects with different stochastic and chaotic properties. Among the most known are Shannon entropy [18], sample entropy (SampEn) [19], permutation entropy (PermEn) [20], fuzzy entropy (FuzzyEn) [58], cosine similarity entropy [59], phase entropy [60], singular value decomposition entropy (SvdEn) [61]. The search for new chaos estimation algorithms includes, for example, Neural Network Entropy (NNetEn) [62–64], which is based on the classification of special datasets in relation to the entropy of the time series recorded in the reservoir of the neural network. NNetEn does not take into account probability distribution functions.

FuzzyEn method for estimating the chaos, used in this study, has a number of adjustable parameters, such as a high-calculation speed, and good sensitivity when using a short time series.

The possibility of approximating entropy using machine learning regression methods was demonstrated by Velichko et al. [65]. The entropies of SvdEn, PermEn, SampEn and NNetEn were used in the study, and the gradient boosting algorithm was recognized as the best regression method. The results of the study were used to accelerate the calculation of 2D image entropy in remote sensing.

Among the neurodynamic models with chaotic dynamics, chaotic 3D models of Hodgkins–Huxley [66], Hindmarsh–Rose [67], modified 2D models of Fitzhugh–Nagumo [68] and Izhikevich [69] can be distinguished.

The Hindmarsh–Rose (HR) system is one of the popular spike models in describing real biological neurons. It appeared as an analysis of the Hodgkin–Huxley equations, which represented a 2D analog of the Fitzhugh–Nagumo neuron, and allowed observation of long intervals between spike sequences [70]. Later, an additional equation was added to the system of equations [8].

One of the applications of the HR model is to simulate the interaction of real neurons in various network architectures. In [71], the influence of the coupling strength and driving current on the behavior of HR neurons in a ring network is studied. In another network with the "all-with-all" architecture, the HR model helped to establish that the difference between the incoming and outgoing current plays a decisive role in the activity of neurons, and this finding can be used to explain the mechanism of short-term memory generation [72]. Various synchronization effects in networks of HR neurons were investigated [9,73,74]. In addition to the usual (tonic) oscillation mode, the HR model describes the burst activity of neurons when tonic oscillations alternate with rest states. In addition, in the HR model, chaos is observed at the moment of transition between simple tonic and bursting modes, and can be explained by the theory of homoclinic bifurcations of co-dimension two, which are typical for fast–slow systems [10]. A feature of such bifurcations is the doubling of the periods of both periodic and homoclinic orbits [75].

In this study, we present two models of the chaos sensor. The first model estimates the irregularity of time series through the calculation of FuzzyEn (Sensor on FuzzyEn—(SFU)). The second model estimates (approximates) the degree of chaos by a multilayer perceptron (Sensor on Perceptron—(SPE)), trained on the results of the SFU model. The goal is to create the SPE model, which, the results, after training the perceptron, would match with the SFU model on $R^2$ metric in the best possible way. The time series datasets for training and testing were created using the Hindmarsh–Rose spike model and the SFU model. The study reviews the following questions. What is the length of the time series that must be fed to the sensor input in order to calculate the entropy with a given accuracy for the SFU model? What are the optimal FuzzyEn parameters for maximum sensitivity of SFU and SPE sensors? What perceptron parameters are optimal for approximating the entropy function? How does the accuracy of the SPE model approximation depend on the time series normalization method? What is the principle of approximating entropy using a perceptron?

The contributions of the study include:

- Optimal FuzzyEn parameters were determined, allowing to reach maximum sensitivity of the chaos sensor for SFU and SPE models;
- Time series datasets for training and testing the perceptron model based on the Hindmarsh–Rose spike model and FuzzyEn were created.
- A bio-inspired model of a chaos sensor based on a multilayer perceptron, approximating FuzzyEn, is proposed.
- The proposed perceptron model with 1 neuron in the hidden layer reaches high degree of similarity between SFU and SPE models with an accuracy in the range $R^2 \sim 0.5 \div 0.8$, depending on the combination of datasets.
- The proposed perceptron model with 50 neurons in the hidden layer reaches an extremely high degree of similarity between SFU and SPE models with an accuracy of $R^2 \sim 0.9$.
- An example of using the chaos sensor on spike train of action potentials recordings from the L5 dorsal rootlet of rat is given.

The rest of the paper is organized as follows. In Section 2, the perceptron model, techniques for modeling the Hindmarsh–Rose system, generating time series, creating datasets, calculating and approximating entropy are described. Section 3 is dedicated to numerical results for SFU and SPE models. Section 4 discusses research results and outlines directions for future research. The conclusion is provided in Section 5.

## 2. Materials and Methods

### 2.1. Multilayer Perceptron Neural Network Model

The model of the perceptron used as a chaos sensor is shown in Figure 2. The SPE model has one hidden layer, and the output is represented by one neuron with a linear activation function $f(z) = z$, where $z$ is the weighted sum at the input of the neuron. The number of inputs corresponds to the number of elements of the time series $NL$. In this study, we consider a model with $NL = 50$. The number of neurons in the hidden layer $NH$ has three gradations $NH = 1$, 50 and 150 with a sigmoid activation function $f(z) = 1/(1 + \exp(-z))$. Figure 2a shows a perceptron model with one neuron in the hidden layer $NH = 1$, and Figure 2b reflects a model with $NH = 50$ neurons in the hidden layer.

A linear activation function is used on the output neuron, as this approach is most common in regression models, and is effective for approximating the entropy of a time series.

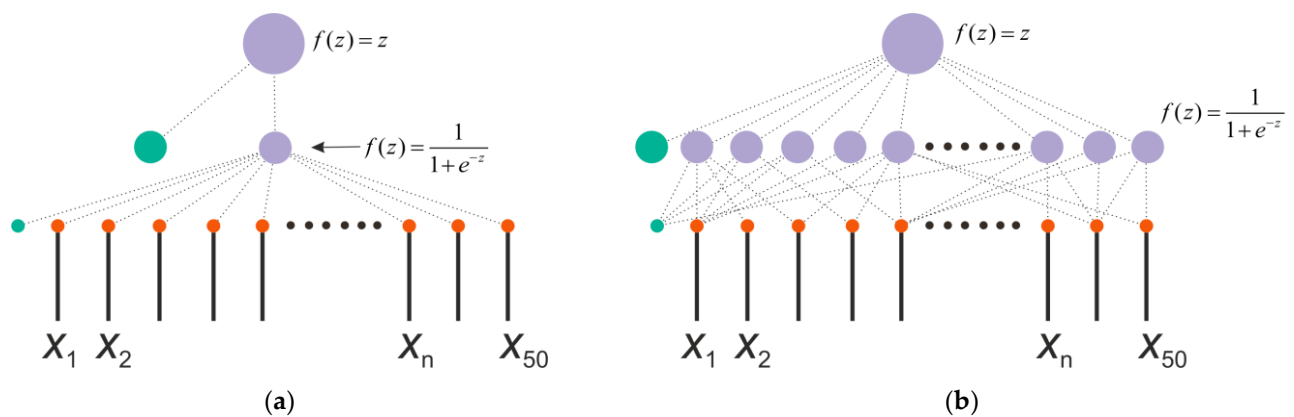

**Figure 2.** Perceptron model with the number of neurons in the hidden layer $NH = 1$ (**a**) and $NH = 50$ (**b**). The figure shows the structure of a perceptron used as a chaos sensor model with a different number of neurons in the hidden layer and one output neuron. The input is a time series $x_n$. The figure shows the activation functions of neurons.

The complete MLP models with weight distribution are presented in Supplementary Materials (Model_1, Model_2).

### 2.2. Modeling the Hindmarsh-Rose System

The Hindmarsh–Rose (HR) system, which is one of the most popular spike models, has the form [8]:

$$\begin{cases} \dfrac{\partial X}{\partial t} = Y + 3X^2 - X^3 - Z + I_{ex} \\ \dfrac{\partial Y}{\partial t} = 1 - 5X^2 - Y \\ \dfrac{\partial Z}{\partial t} = r[4(X + 8/5) - Z] \end{cases}, \quad (1)$$

where $X$ is the membrane potential of the cell, $Y$ and $Z$ are the concentration of sodium and potassium ions, $I_{ex}$ is the external current (stimulus), $r$ is a small parameter characterizing the slow potassium current and, $t$ is time.

Figure 3a represents typical chaotic oscillograms of the variables of the HR model and the distances between the spikes for $X(t)$ (Figure 3b). Figure 4 shows a bifurcation diagram of the spike distances of the variable $X(t)$ with a change in the small parameter $r$ for two values of the external current $I_{ex}$ = 3.25 (Figure 4a) and $I_{ex}$ = 3.35 (Figure 4b).

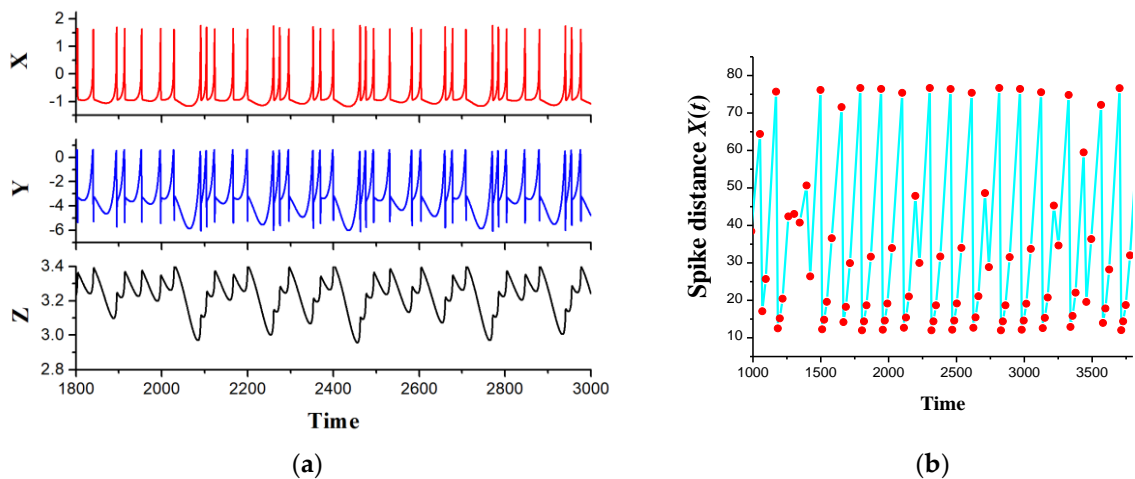

(**a**) (**b**)

**Figure 3.** (**a**) Oscillograms of HR model variables calculated from equations (1). (**b**) Oscillogram of distances between spikes of cell membrane potential $X(t)$. Model parameter $r$ = 0.0055.

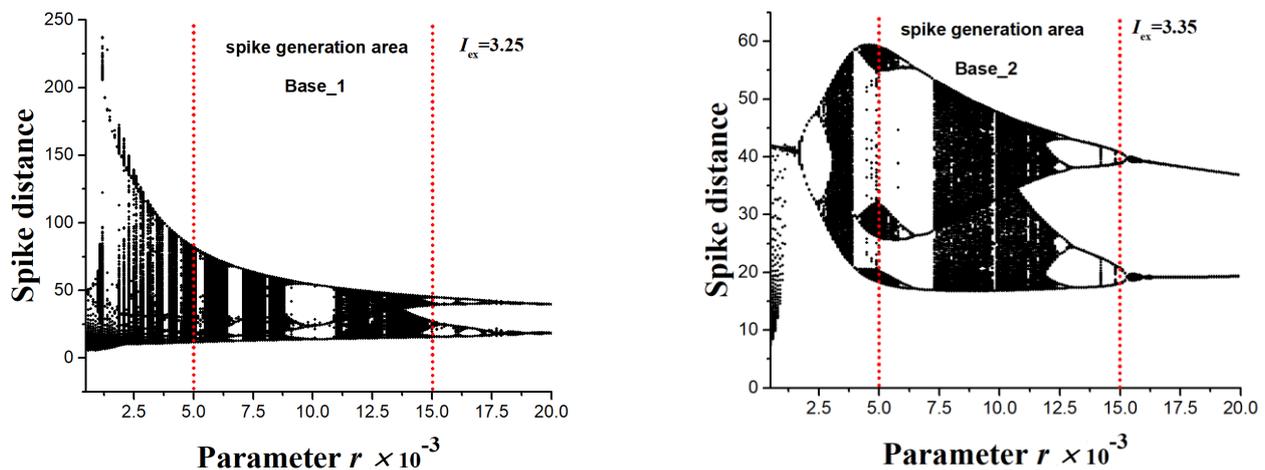

(**a**) (**b**)

**Figure 4.** Bifurcation diagrams of distances between spikes of the HR model for the variable *X*(*t*) with a change in the parameter *r* for two values of the external current (**a**) $I_{ex}$ = 3.25 and (**b**) $I_{ex}$ = 3.35. The region of spike generation for the Base_1 (**a**) and Base_2 (**b**) datasets used for entropy calculation and perceptron training is shown. The diagrams demonstrate the variety of chaotic and regular dynamics of oscillations with varying control parameters *r* and $I_{ex}$.

The dynamics of the HR model are well studied [67,8] and present an alternation of regular and chaotic modes. This behavior is helpful for creating datasets for training the perceptron model as a chaos sensor.

## 2.3. Method for Generating Time Series of Various Lengths

The method of generating time series of various lengths was used to generate datasets and to study the dependence of the accuracy of the chaos sensor model on the length of the time series. Series sets of length *NL* from 10 to 100 elements with a step of 10 were generated. Each set contained time series for both chaotic and regular modes of spike oscillograms; the mode is controlled by the parameter *r* of Equation (1). The range of *r* was {$5×10^{-3} \div 1.5×10^{-2}$}, as shown in Figure 4.

For a given value of *r*, the generated oscillogram was used to obtain a long time series with 500 elements. Based on one such long series, 100 short time series of the same length *NL* were generated by shifting the window of a short time series along the main series with a step *S* (Figure 5). The step size was *S* = 4. Due to this method, the short series differed from each other in the dataset, even they belonged to the same state of the HR oscillator. Consequently, the series were generated in a wide variety and reflected the data flow to the chaos sensor in the continuous measurement mode.

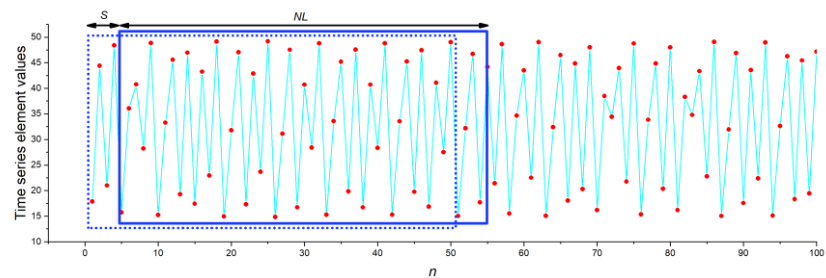

**Figure 5.** The principle of generating sets of short time series within one long time series. The blue frame, which is shifted with step *S*, indicates the short series window with a length of *NL*.

## 2.4. Datasets for Training and Testing the Perceptron Using HR System

To train and test the perceptron, we used two different datasets, Base_1 and Base_2, which were obtained by modeling the HR system, as well as their combination Base_1_2. Datasets Base_1 and Base_2 have different time series sets because of differences in values of the external current, $I_{ex}$, as noticeable in the buffer diagrams (Figure 4).

- The Base_1 dataset consisted of 10,000 time series with a length of *NL* = 50, which were obtained by modeling the HR system with $I_{ex}$ = 3.25 (Figure 4a). The range of *r* was {$5×10^{-3} \div 1.5×10^{-2}$} and it was divided into 100 values. One-hundred short time series for each *r* were generated according to the algorithm from Section 2.3. The target value of the entropy of all series was estimated through FuzzyEn using the method from Section 2.6.
- The Base_2 dataset consisted of 10,000 time series with a length of *NL* = 50, which were obtained by modeling the HR system with $I_{ex}$ = 3.35 (Figure 4a). The range of *r* was {$5×10^{-3} \div 1.5×10^{-2}$} and it was divided into 100 values. The generation of 100 short time series for each *r* was performed according to the algorithm from Section 2.3. The target value of the entropy of all series was estimated through FuzzyEn using the method from Section 2.6.
- Dataset Base_1_2 was a combination of Base_1 and Base_2.

The main characteristics of the datasets are presented in Table 1. *Mean* is the main characteristic and is the average value for all elements of the time series in the dataset. In addition, *Mean50_min* is the minimum value of the average value of the time series, consisting of 50 elements; *Mean50_max* is the maximum value of the average value of the time series, consisting of 50 elements; *Min_X* is the minimum value of all elements in the dataset; *Max_X* is the maximum value of all elements in the dataset. Dataset files are available in the Supplementary Materials section.

*Min_X* and *Max_X* of the two datasets (Base_1 and Base_2) differ significantly. This difference was created by us intentionally in order to test the universality of the created perceptron models as a chaos sensor, operating with different time series of spike train in neurodynamic systems. The *r* range {$5 \times 10^{-3} \div 1.5 \times 10^{-2}$} and two current values, $I_{ex}$ = 3.25 and $I_{ex}$ = 3.35, allowed us to create extensive datasets containing chaotic and ordered modes of operation of the HR system. This setup makes it possible to study the chaos sensor model in a wide range of time series dynamics.

**Table 1.** Characteristics of datasets for training and testing the perceptron.

| Datasets | *Mean* | *Mean50_min* | *Mean50_max* | *Min_X* | *Max_X* |
|---|---|---|---|---|---|
| Base_1 | 32.26571 | 30.09417 | 36.30281 | 11.686 | 81.764 |
| Base_2 | 32.28749 | 29.70656 | 36.25960 | 16.744 | 59.124 |
| Base_1_2 | 32.27660 | 29.70656 | 36.30281 | 11.686 | 81.764 |

Time series from datasets were used to train and test the perceptron. Preliminary normalization of series was performed by subtracting the *Mean* value from the values of the series elements (Table 1).

*2.5. Datasets for Training and Testing the Perceptron Using Experimental Data*

To train and test the perceptron on experimental data, we compiled datasets based on open data of action potentials recorded from the L5 dorsal rootlet of rat [41,76]. The original dataset contains 20 recordings in total: 10 recordings were made with the animal at rest and 10 recordings were performed during stimulation. Each recording contains five channels. To compile a dataset of spike periods of action potentials, we only used the first channel. To extract the times of the spikes, we used the software tool Peak Analyzer in Origin Pro with the following parameters: goal = find peaks, base line mode = none (Y = 0), enable auto find = true, peak filtering by height 20%, smoothing window size = 0, direction = positive, method = local maximum, local points = 2. Before searching for peaks, the oscillogram was filtered with a low-pass filter with a frequency of 10 kHz. Examples of spike train and peak location are presented in Figure 6a when the animal is resting, and in Figure 6b when the animal is stimulated. Stimulation is characterized by a higher frequency of peaks.

The training set (Base_1_exp) consisted of time series of periods between spikes, composed of 18 records, divided into short time series according to the method in Section 2.3 at *S* = 1 and *NL* = 50. We used 9 recordings at rest (824 time series in total) and 9 recordings in the stimulation state (1593 time series in total). Basic statistical parameters of Base_1_exp are *Mean* = 0.0013, *Min_X* = 0.0001, *Max_X* = 0.0052. The test set (Base_2_exp) consisted of time series of 50 elements of periods between spikes, 1 recording at rest state (49 time series in total) and 1 recording in the stimulation state (63 time series in total). Basic statistical parameters of Base_2_exp are *Mean* = 0.0015, *Min_X* = 0.0001, *Max_X* = 0.0088. Dataset files are available in the Supplementary Materials section.

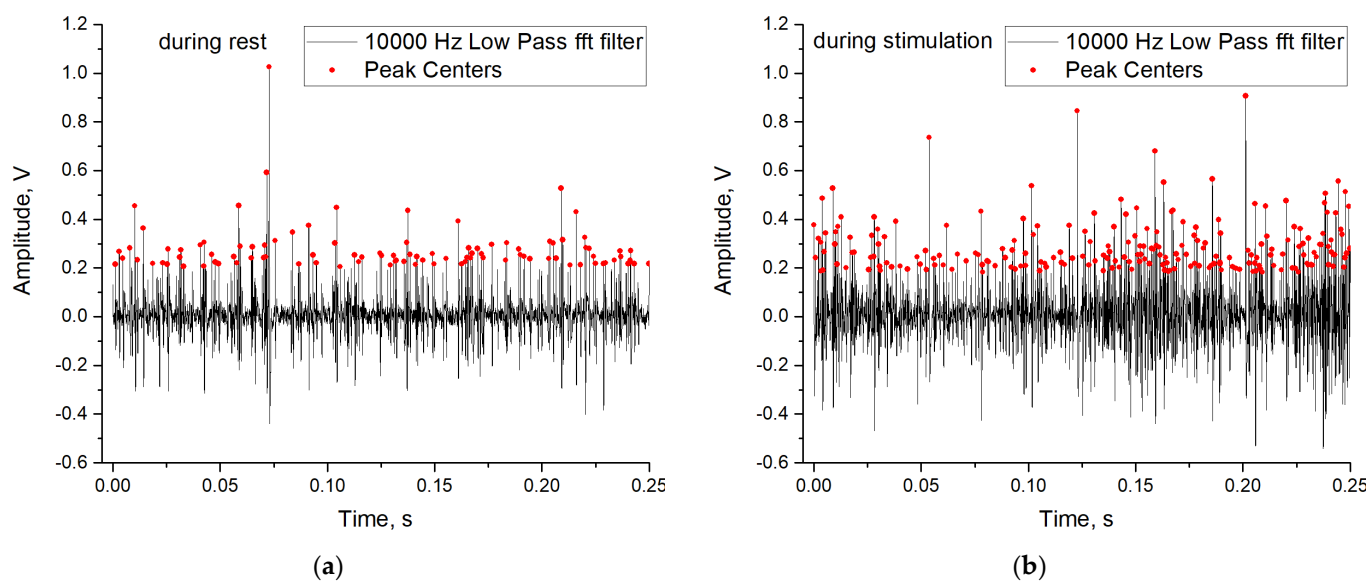

**Figure 6.** Examples of spike train and peak localization during rest (**a**) and during stimulation (**b**).

The entropy distribution (SFU model) for the training set depending on time series number (Base_1_exp) is shown in Figure 7a, and distribution for the test base (Base_2_exp) is presented in Figure 7b. The entropy has a higher value in the rest phase of the animal ($i ≤ 824$) with the average value for Base_1_exp of ~4.056; in the stimulation phase ($i > 824$), the average entropy is lower and equals ~3.481. In the stimulation series, a wide range of entropy values are observed because the stimulation was performed at certain time intervals, not over the entire time interval.

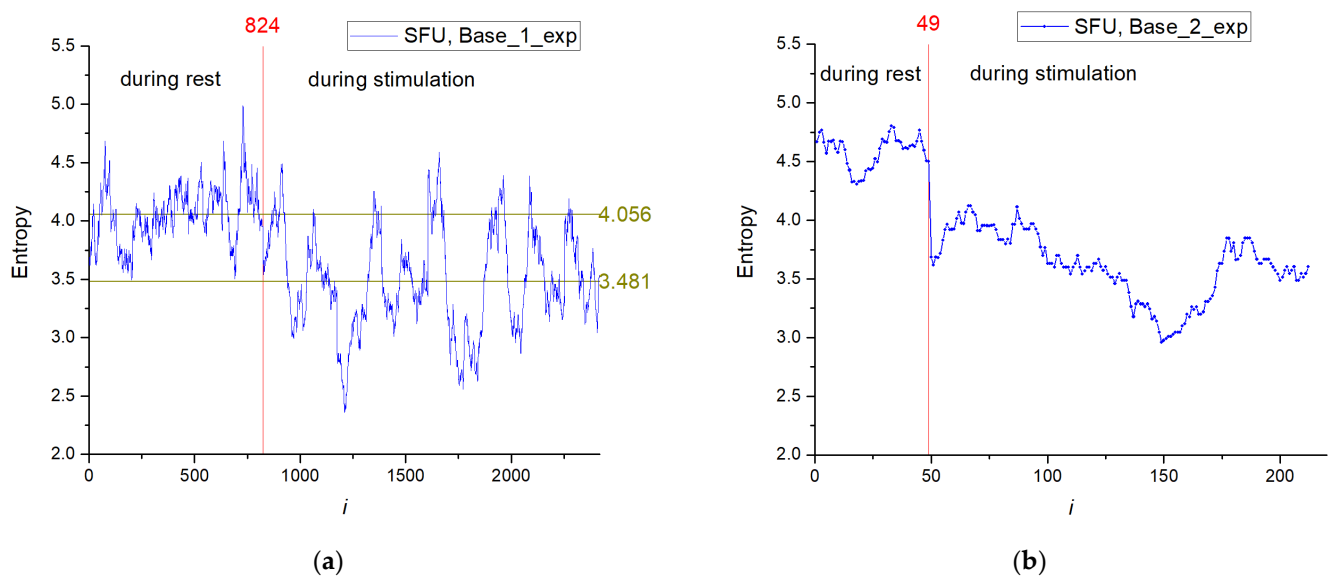

**Figure 7.** The entropy distribution depending on the element number in Base_1_exp (**a**) and Base_2_exp (**b**).

*2.6. Entropy Calculation Method*

We used FuzzyEn as the main method for assessing chaos, as it has good sensitivity when using short time series, a number of adjustable parameters, and a high-calculation speed.

FuzzyEn [77] is calculated by measuring the conditional probability of the fact that if vectors (segments of a time series) of length *m* are similar, taking into account the tolerance $r_1$, then vectors of length $m + 1$ will be similar with the same tolerance $r_1$. The similarity metric between vectors is a fuzzy function calculated based on the distance between a pair of compared vectors.

For a time series vector $T$ of length $N$, it is possible to compose $N − m + 1$ vectors $X_i^m$ of length $m$, consisting of successive segments of the original series $T$, from which the average value of the elements of the vector $T_i^{avg}$ was subtracted:

$$X_i^m = \{x_i, x_{i+1}, ..., x_{i+m-1}\} = \{T_i, T_{i+1}, ..., T_{i+m-1}\} - T_i^{avg} \tag{2}$$

where $i = 1…N − m + 1$, and $T_i^{avg}$ is calculated as:

$$T_i^{avg} = \frac{1}{m} \sum_{j=0}^{m-1} T_{i+j} \tag{3}$$

For a vector $X_i^m$, we can define a distance $d_{ij}^m$ between this vector and vector $X_j^m$. The distance is equal to the maximum absolute difference between the components of the vector:

$$d_{ij}^m = \max_{k \in (0, m-1)} |x_{i+k} - x_{j+k}| \tag{4}$$

The similarity between vectors is determined by a fuzzy function $D_{ij}^m$:

$$D_{ij}^m = \exp\left(-\frac{(d_{ij}^m)^{r_2}}{r_1}\right) \tag{5}$$

The entropy value FuzzyEn is calculated based on the values of the average similarity of vectors and for a finite series $T$ can be expressed as:

$$FuzzyEn(m, r_1, r_2) = \ln(\phi^m(r_1, r_2)) - \ln(\phi^{m+1}(r_1, r_2)) \tag{6}$$

where function $\varphi^m$ is expressed as:

$$\phi^m(r_1, r_2) = \frac{1}{N-m} \sum_i^{N-m} \left( \frac{1}{N-m-1} \sum_{j=1, j \neq i}^{N-m} D_{ij}^m \right) \tag{7}$$

The variation range during parameter optimization was for the $m$ from 1 to 3, for $r_2$ from 1 to 5, for $r_1$ from (0.005÷0.4) × $std$, where $std$ is the standard deviation of the elements of the time series.

To calculate the entropy, the python library EntropyHub was used [78].

Optimization method indicates the parameters FuzzyEn ($m = 1$, $r_2 = 1$, $r_1 = 0.01 \times std$) provide the highest sensitivity for time series lengths from 10 to 100 elements.

*2.7. Perceptron Training and Testing Method*

The goal of the study is to create the SPE model, which, the results, after training the perceptron, would match with the SFU model on $R^2$ metric [79] in the best possible way.

To achieve this, it is necessary to train the perceptron to approximate FuzzyEn values when applying all 50 elements of the time series to the input. The datasets Base_1, Base_2, Base_1_2, Base_1_exp, Base_2_exp and the scikit-learn python library were used to train and test the perceptron. Two-layer models were used with the number of neurons in the hidden layer being 1, 50, and 150. The selection of model hyperparameters and accuracy assessment were carried out using the K-fold cross-validation method (K = 10). In this method, the dataset is divided into K-folds. The process of training and testing the model is performed K times, each time one fold is used for testing, and all other data make up the training set. As a metric for assessing the accuracy of the model, we used the coefficient of determination $R^2$, MAPE and RMSE.

To evaluate the model's universality, the method of training the model on one dataset and testing the model on another dataset was used.

*2.8. Sensor Characteristics*

As the main characteristics of SFU and SPE sensors, the following were identified:

- $En_{av}$(chaos): average value of entropy over five chaotic series, at $r$ = 0.0056, 0.0076, 0.0082, 0.0119, 0.0141 in HR model; Averaging within each series was performed over 100 short series.
- $En_{av}$(order): average entropy value over five regular series, at $r$ = 0.0068, 0.0070, 0.0099, 0.0105, 0.0108 in HR model; Averaging within each series was performed over 100 short series.
- $EnR = En_{av}$(chaos) − $En_{av}$(order): range of entropy change at the output of the sensor.
- $Std\_En$(chaos): entropy mean square deviation over five chaotic series;
- $Std\_En$(order): entropy mean square deviation over five regular series;
- $EnSens = EnR/ Std\_En$(chaos): chaos sensor sensitivity.
- $EnErr = (Std\_En$(chaos)$/EnR) \times 100\%$: relative entropy measurement error in percent.

The presented characteristics were calculated at the current $I_{ex}$ = 3.25.

The range of entropy change $EnR$ shows how large the output range of the sensor is when the signal dynamics changes from order to chaos. The sensitivity of the $EnSens$ sensor shows how many times the mean square deviation is less than the range of entropy change. The relative measurement error ($EnErr$) shows the uncertainty of the output entropy value in percent; it is inversely proportional to the sensitivity. These characteristics may have different values in SFU and SPE sensor models.

Before calculating the dependence of sensor characteristics on $NL$, it is necessary to determine the optimal FuzzyEn entropy parameters, at which the sensitivity of the SFU model has maximum values for $NL$ = 10÷100. The optimization method found the parameters FuzzyEn ($m$ = 1, $r_2$ = 1, $r_1$ = 0.01 × $std$) that meet these requirements.

## 3. Results

*3.1. Dependence of Sensor Characteristics on the Series Length in the SFU Model.*

In this study, we present two models of the chaos sensor. The first model evaluates the irregularity of the time series through a FuzzyEn calculation called the SFU model (sensor). The second model estimates (approximates) the degree of chaos by a multilayer perceptron called the SPE model (sensor); this model is pre-trained on the results of the SFU model. The concept of the chaos sensor developed in the paper implies that a time series of a certain length $NL$ is the input, and the entropy value is calculated at the output. As an SPE sensor, we used a multilayer perceptron with the number of inputs equivalent to the length of the series $NL$. To answer the question of how the time series length affects the sensor characteristics, we generated sets of time series of various lengths $NL$ from 10 to 100 elements with a step of 10 according to the method from Section 2.3.

Figure 8a shows the dependences of $En_{av}$(chaos), $En_{av}$(order), and $EnR$ on the length of the $NL$ series for the SFU model. At $NL$ > 40, the values of the characteristics saturate; it indicates the independence of the entropy values from the length of the series when this threshold is exceeded, and indicates the stability of the calculation of FuzzyEn on short time series with optimal parameters. As the series length decreases from $NL$ = 40 to $NL$ = 10, a significant increase in $En_{av}$ (chaos) and $EnR$ is observed. A decrease in the length of a series can lead to both an increase and a decrease in entropy, depending on the entropy parameters, as shown, for example, for NNetEn [63]. The range of entropy change is of the order of $EnR$~3; it is more than twice as large as $En_{av}$ (order) and reliably records the chaotic behavior.

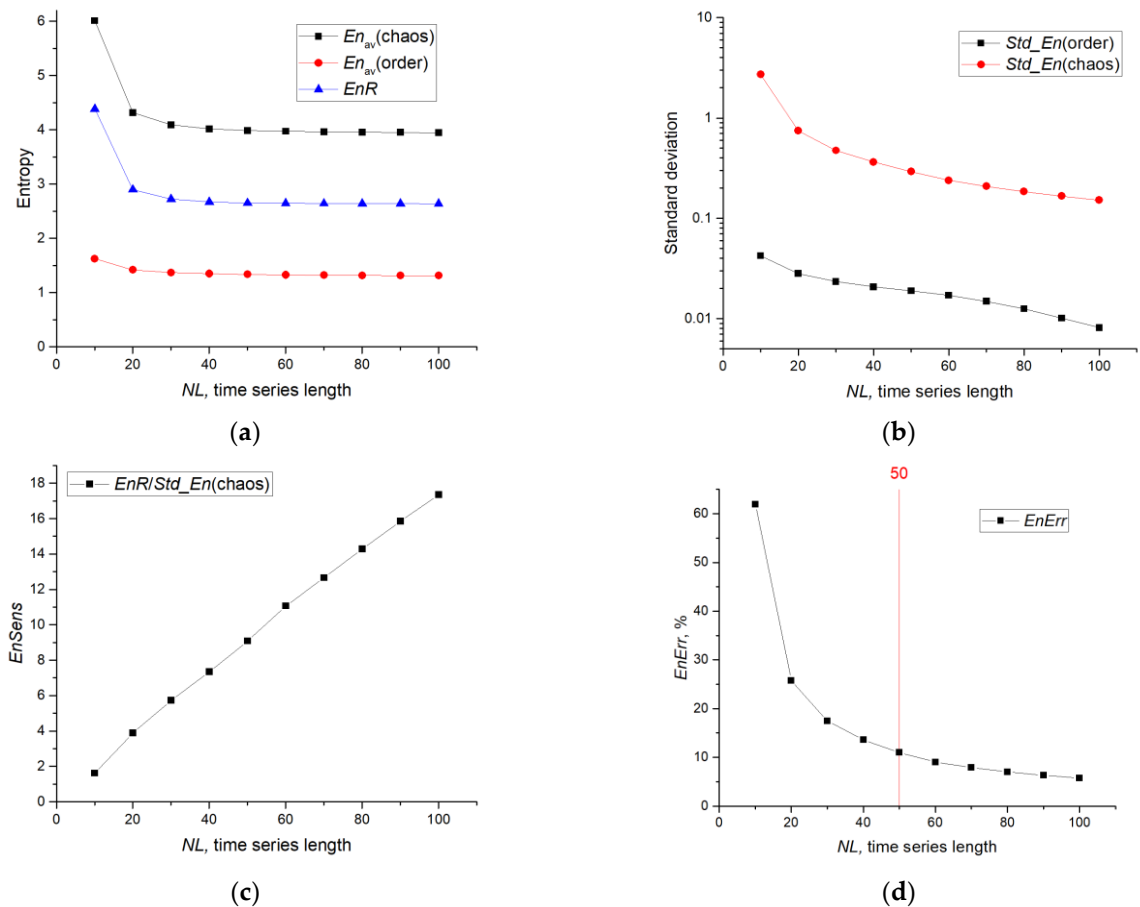

**Figure 8.** Dependence of chaos sensor characteristics on the length of the series in the SFU model (Sensor on FuzzyEn). The graphs show (**a**) *En*av(chaos) and *En*av (order), (**b**) *Std_En*(chaos) and *Std_En*(order), (**c**) *EnSens* sensitivity, (**d**) *EnErr* relative error.

Figure 8b shows the dependencies of *Std_En*(chaos) and *Std_En*(order) on *NL*. *Std_En*(chaos) exceeds *Std_En*(order) by more than an order of magnitude, and both characteristics decrease rapidly with increasing *NL*. At *NL* = 50, *Std_En*(chaos) ~ 0.3, it is approximately 10 times less than *EnR*. Figure 8c shows the dependence of the sensitivity of the *EnSens* sensor on the length *NL* series. *EnSens* increases linearly with the length of the time series, and it indicates a better measurement of chaos on long series. The increase in *EnSens* is mainly associated with a decrease in the standard deviation *Std_En*(chaos), since *EnR* almost does not change from *NL*. The relative entropy measurement error *EnErr* as a function of *NL* is shown in Figure 8d. The maximum measurement error *EnErr* ~ 60% is observed at the minimum time series length *NL* = 10 and quickly decreases to *EnErr* ~ 11% at *NL* = 50. The measurement error *EnErr* ≤ 11% can be considered acceptable in practice, and the sensor requires a time series with the length of *NL* ≥ 50.

*3.2. Results of Entropy Approximation in Perceptron SPE Model*

To train the perceptron, the series length *NL* = 50 was chosen based on the results of the previous section, where the error *EnErr* ~ 11% was demonstrated for the SFU model.

Table 2 presents the estimation results of the determination coefficient $R^2$ for perceptron models with the number of neurons in the hidden layer *NH* = 1, 50 and 150. Results for RMSE and MAPE metrics are presented in Tables A1 and A2 (Appendix A). Preliminary normalization was not performed. Various combinations of training and test datasets were evaluated, submitting time series without normalization. For *NH* = 1, the models poorly approximated the entropy in all cases. At *NH* = 50 and *NH* = 150, the approximation had high values of $R^2 \sim 0.8 \div 0.88$ when using the cross-validation method within the baseline. The model with a large number of neurons in the hidden layer $NH$ = 150 showed the best results. Cross-validation within the combined set Base_1_2 showed high results, although separate training on one set and testing on another set demonstrated low $R^2$ values. It indicates that the model is being over trained on a separate set and is not suitable for working with another set. In general, the approach without normalization can be applied in practice if the training set contains all possible combinations of time series, as in the case of Base_1_2.

**Table 2.** Estimation of approximation accuracy by $R^2$ metric without the normalization stage.

| Number of Neurons in the Hidden Layer, *NH* | Cross-Validation Accuracy, $R^2$ | | | Approximation Accuracy When Training on One Set and Testing on Another Set, $R^2$ | |
|---|---|---|---|---|---|
| | Base_1 | Base_2 | Base_1_2 | Base_1-Training Base_2-Testing | Base_1-Testing Base_2-Training |
| 1 | −0.001 | −0.001 | −0.001 | −0.013 | −0.047 |
| 50 | 0.827 | 0.868 | 0.810 | 0.030 | −0.084 |
| 150 | 0.846 | 0.888 | 0.842 | 0.159 | −0.831 |

Table 3 presents the estimation results of the coefficient of determination $R^2$ for perceptron models with the number of neurons in the hidden layer $NH$ = 1, 50 and 150, with preliminary normalization of the time series. Results for RMSE and MAPE metrics are presented in Tables A3 and A4 (Appendix A). Normalization was performed by subtracting the value of the average *Mean* (Table 1) from the values of the elements within the series.

**Table 3.** Approximation accuracy estimation by metric $R^2$ using the normalization stage.

| Number of Neurons in the Hidden Layer, *NH* | Cross-Validation Accuracy, $R^2$ | | | Approximation Accuracy When Training on One Set and Testing on Another Set, $R^2$ | |
|---|---|---|---|---|---|
| | Base_1 | Base_2 | Base_1_2 | Base_1-Training Base_2-Testing | Base_1-Testing Base_2-Training |
| 1 | 0.812 | 0.617 | 0.498 | 0.128 | 0.326 |
| 50 | 0.904 | 0.928 | 0.885 | −0.180 | 0.655 |
| 150 | 0.912 | 0.937 | 0.897 | −0.729 | 0.568 |

Models with one neuron in the hidden layer ($NH$ = 1) had high values of $R^2 \sim 0.5 \div 0.8$ in the cross-validation mode. The largest $R^2 \sim 0.8$ corresponded to Base_1, and the smallest $R^2 \sim 0.5$ corresponded to the combined Base_1_2. Even a single neuron in the hidden layer approximates entropy well, although the accuracy of the model decreases with an increase in the variety of the time series when the datasets are combined (Base_1_2). Figure 9a shows the dependence of FuzzyEn for the SFU model on the series number in Base_1 and the dependence predicted by the SPE model with one neuron (Perceptron $NH$ = 1). In addition, the graph shows the average value over 20 measurements for the SPE model (Averaging over 20). The perceptron model predicts the FuzzyEn entropy values well, repeating the regions with high- and low-entropy values. To confirm the correctness of the calculation of the SFU and SPE models, the insertion of the bifurcation diagram of the HR model corresponding to the Base_1 numbering is provided. The areas of chaos and order correspond to each other. The average value repeats the entropy dependence even better, having less fluctuation. A more detailed study of the physics of the perceptron model with one neuron is discussed in the next section. An increase in the number of neurons in the hidden layer enhances the accuracy of the perceptron SPE model, reaching $R^2 \sim 0.9$ for all datasets in the cross-validation mode. In general, the difference between *NH* = 50 and *NH* = 150 is not significant, and it is acceptable to use *NH* = 50 in practice. Figure 9b shows the entropy curve for the SPE model with *NH* = 50 and Base_1. The perceptron model in this case reproduces FuzzyEn well, and the average value (Averaging over 20) correctly tracks even narrow regions of order and chaos.

The model with normalization performed well when testing and training were conducted on different datasets. With *NH* = 50, training on Base_2 and testing on Base_1 showed $R^2 \sim 0.655$. When the datasets were switched, the approximation deteriorated to $R^2 \sim -0.18$. This suggests that it is better to use Base_2 for training, since it contains more variants present in Base_1, and is more versatile. When combining Base_1_2 bases, cross-validation shows improved values of $R^2 \sim 0.885$. Figure 9c shows the dependences of the entropies for *NH* = 50 and Base_1, in the case of training the model on Base_2. The model track segments with regular time series worsen, but, in general, it estimates FuzzyEn values with acceptable accuracy.

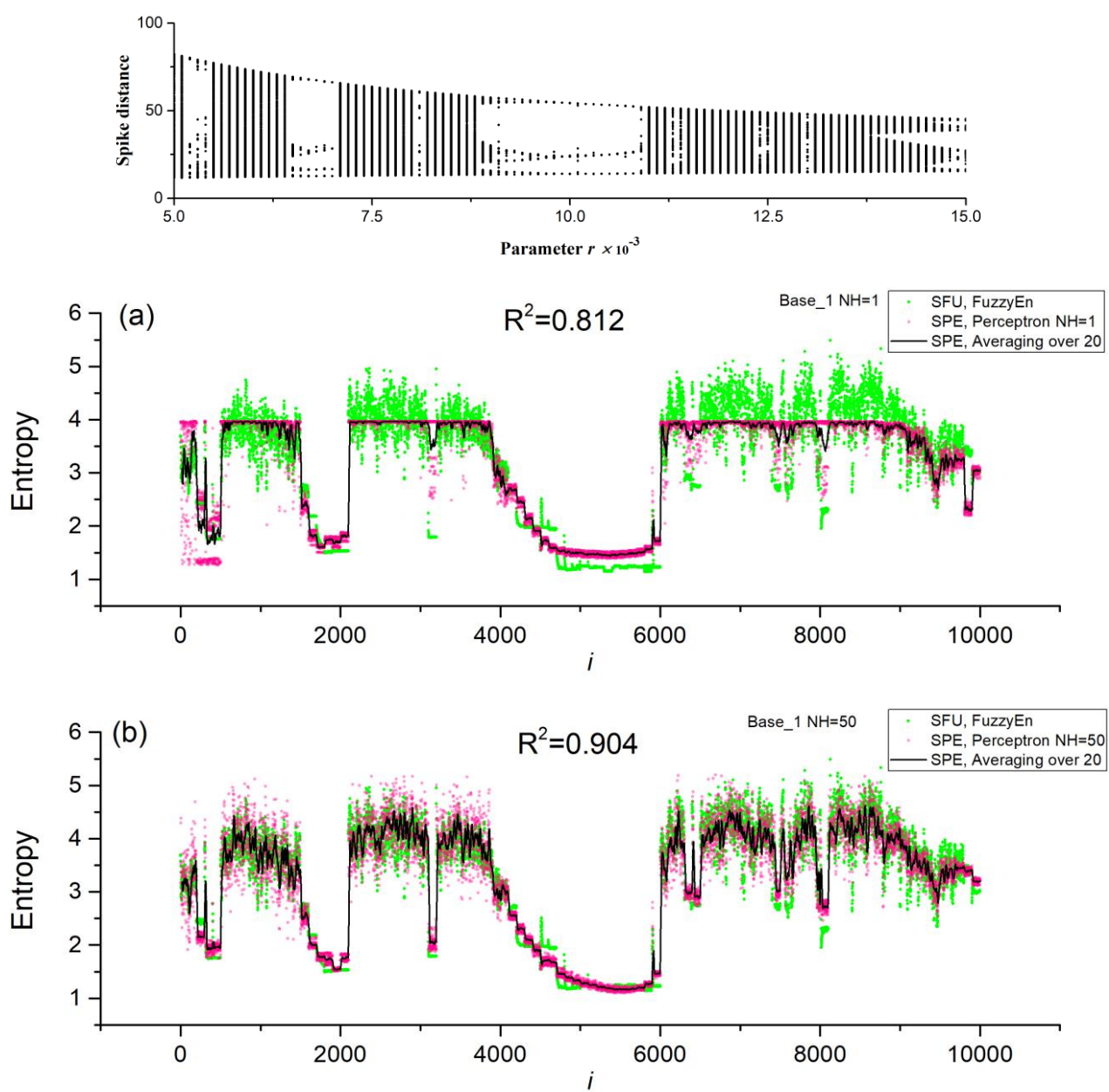

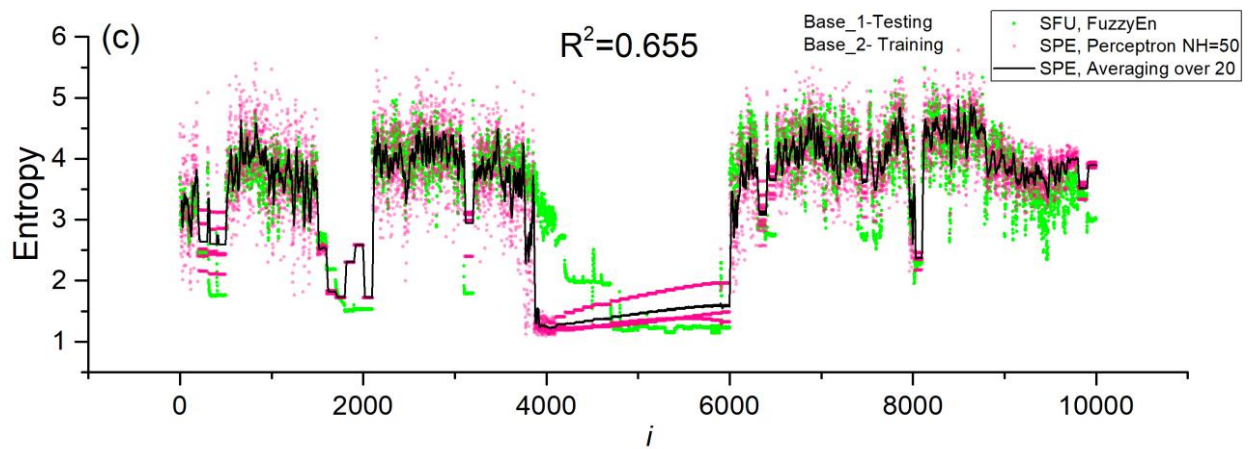

**Figure 9.** Dependences of entropy on the element number in Base_1 for SFU and SPE models, as well as the average value over 20 measurements for SPE (Averaging over 20). Results are shown for models (**a**) Base_1 and *NH* = 1, (**b**) Base_1 and *NH* = 50, (**c**) Base_1—testing, Base_2 -training. The inset to figure (**a**) shows the bifurcation diagram of the HR model corresponding to Base_1.

Figure 10a shows the dependences of the entropies of SFU and SPE sensors for *NH* = 50 and Base_2 after cross-validation. The perceptron model approximates FuzzyEn with an accuracy of $R^2 \sim 0.928$. To confirm the correctness of the calculation of SFU and SPE models, the bifurcation diagram of the HR model corresponding to the Base_2 numbering is inserted. The areas of chaos and order correspond to each other. Figure 10b shows the dependences of the entropies of SFU and SPE sensors for *NH* = 50 and Base_2 in the case of training the model on Base_1. The SPE model poorly tracks segments with regular time series for $i < 2500$, but the model is good at estimating chaotic segments and part of regular segments for $i > 2500$. This leads to the fact that the estimate of the approximation accuracy decreases to $R^2 \sim -0.18$. It can be concluded that it is better to use Base_2 for training, since it contains more options present in Base_1, and is more versatile.

Figure 11 shows the entropy dependences for *NH* = 50 for the combined dataset Base_1_2 after cross-validation. The perceptron model approximates well the FuzzyEn values for both datasets with an accuracy of $R^2 \sim 0.885$. The good accuracy of the approximation allows this model to be used in practice. Table 4 shows the main characteristics of the chaos sensor model at *NL* = 50 for the SFU model, perceptron SPE model ($R^2 \sim 0.928$) and averaging (SPE Averaging over 20). The perceptron model is not inferior to the SFU model in terms of the *EnSens* and *EnErr* parameters, while averaging provides a significant increase in sensitivity and a decrease in the relative measurement error by almost three times to 4%.

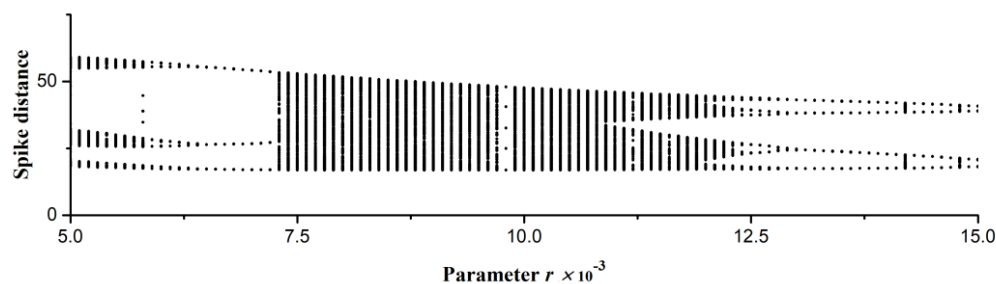

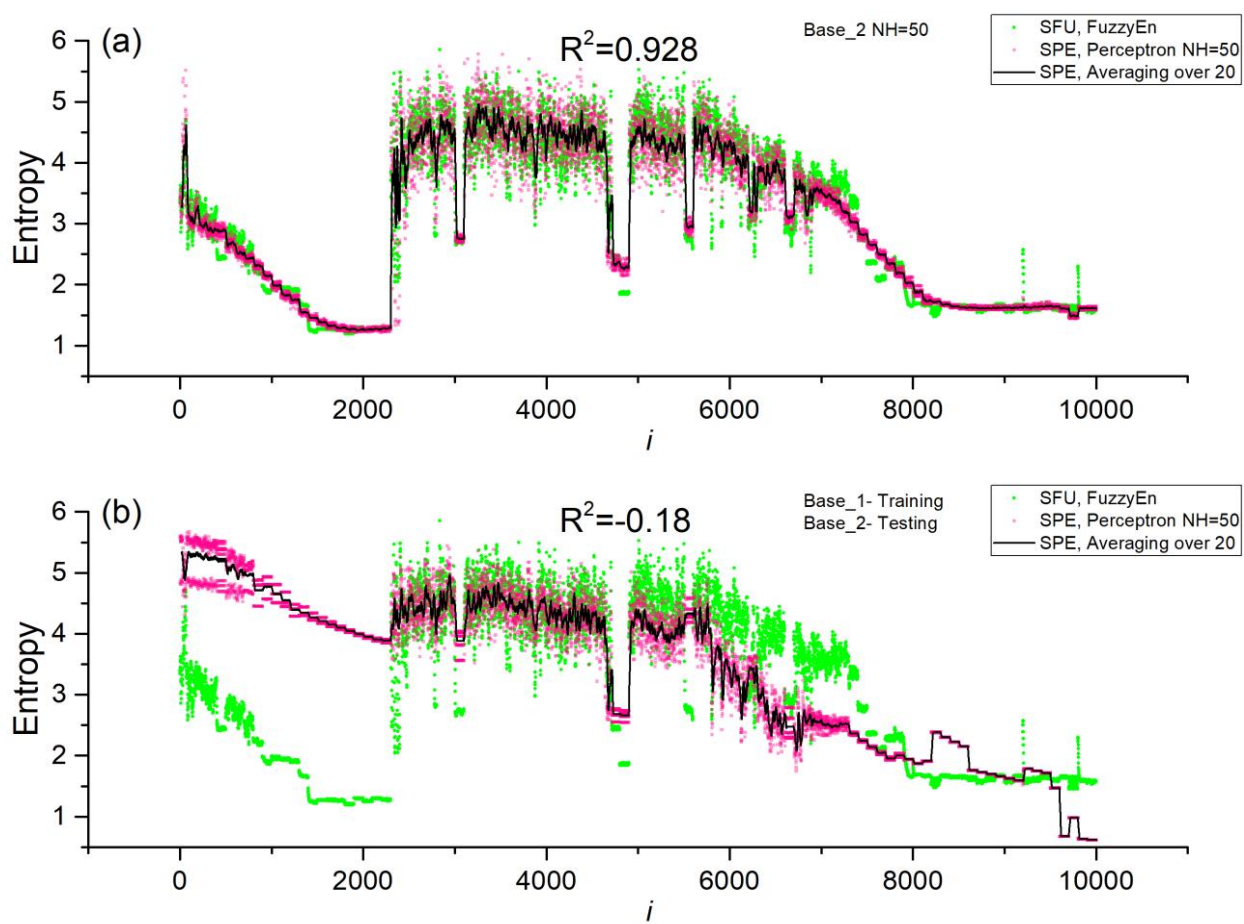

**Figure 10.** Dependences of entropy on the element number in Base_2 for SFU and SPE models, as well as the average value over 20 measurements for SPE (Averaging over 20). Results are shown for models *NH* = 50, (**a**) Base_2, (**b**) Base_2 testing, Base_1 training. The inset to figure (**a**) shows the bifurcation diagram of the HR model corresponding to Base_2.

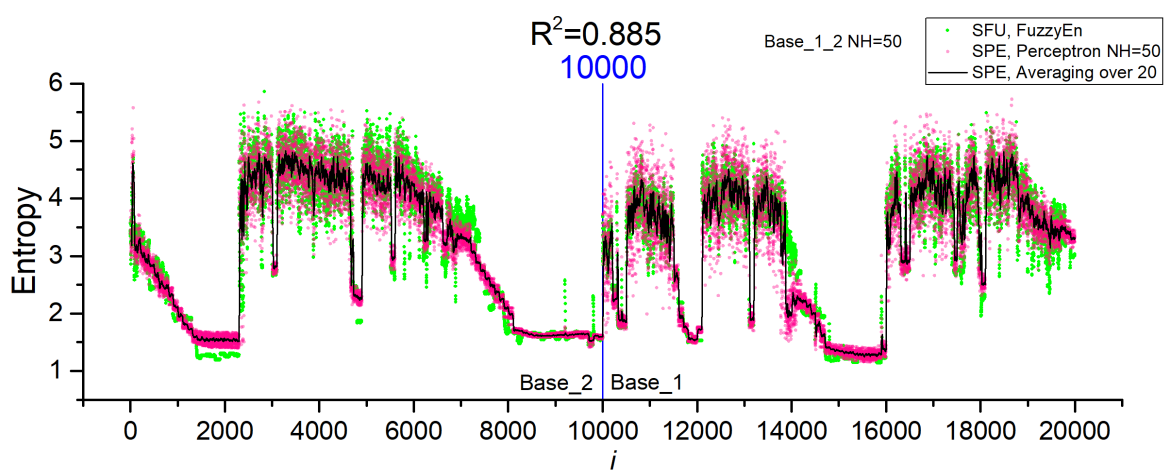

**Figure 11.** Dependences of entropy on the element number in Base_1_2 for SFU and SPE models, as well as the average value over 20 measurements for SPE (Averaging over 20). The results are shown for the *NH* = 50 model for the combined dataset Base_1_2.

**Table 4.** Chaos sensor characteristics at *NL* = 50 for SFU model, perceptron SPE model ($R^2$~0.928) and averaging (SPE Averaging over 20). Perceptron model *NH* = 50 for Base_2 after cross-validation.

|  | $En_{av}$ (Order) | $En_{av}$ (Chaos) | $Std\_En$ (Chaos) | *EnSens* | *EnErr* |
|---|---|---|---|---|---|
| Sensor on FuzzyEn | 1.33 | 3.98 | 0.29 | 9 | 11% |
| Sensor on Perceptron | 1.48 | 4 | 0.28 | 9 | 11% |
| Sensor on Perceptron Averaging over 20 | 1.34 | 4.1 | 0.11 | 25 | 4% |

*3.3. Operating Principles of the Perspectron Sensor Model*

The results of the previous section showed that the use of normalization significantly improves the performance of the SPE chaos sensor. The sensor can work with acceptable accuracy even with one neuron in the hidden layer. To understand the principle of the sensor, let us consider the distribution of weights in the perceptron with one neuron in the hidden layer (Figure 12). The figure shows three options for the distribution of weights for a model based on Base_1 (Figure 12a), Base_2 (Figure 12b) and Base_1_2 (Figure 12c). All three distributions have a common pattern, all weights are positive and have values near the average, indicated by a solid horizontal line in each figure. This regularity suggests replacing all weights with the same values equivalent to the average value. As a result, for Base_1 we obtain the perceptron model shown in Figure 12d. The figure shows the values of all weights. Such a simplified perceptron provides the approximation accuracy as $R^2$ ~ 0.747 for Base_1, $R^2$ ~ 0.236 for Base_2, and $R^2$ ~ 0.447 for Base_1_2. The dependencies of entropy on the element number in Base_1 for the SFU and SPE models are shown in Figure 13. The SPE model roughly approximates the SFU model, however, it traces the main areas of order and chaos reasonably well. Such a simple model performs the averaging of the time series, and the value of the average is then linearly transformed into entropy.

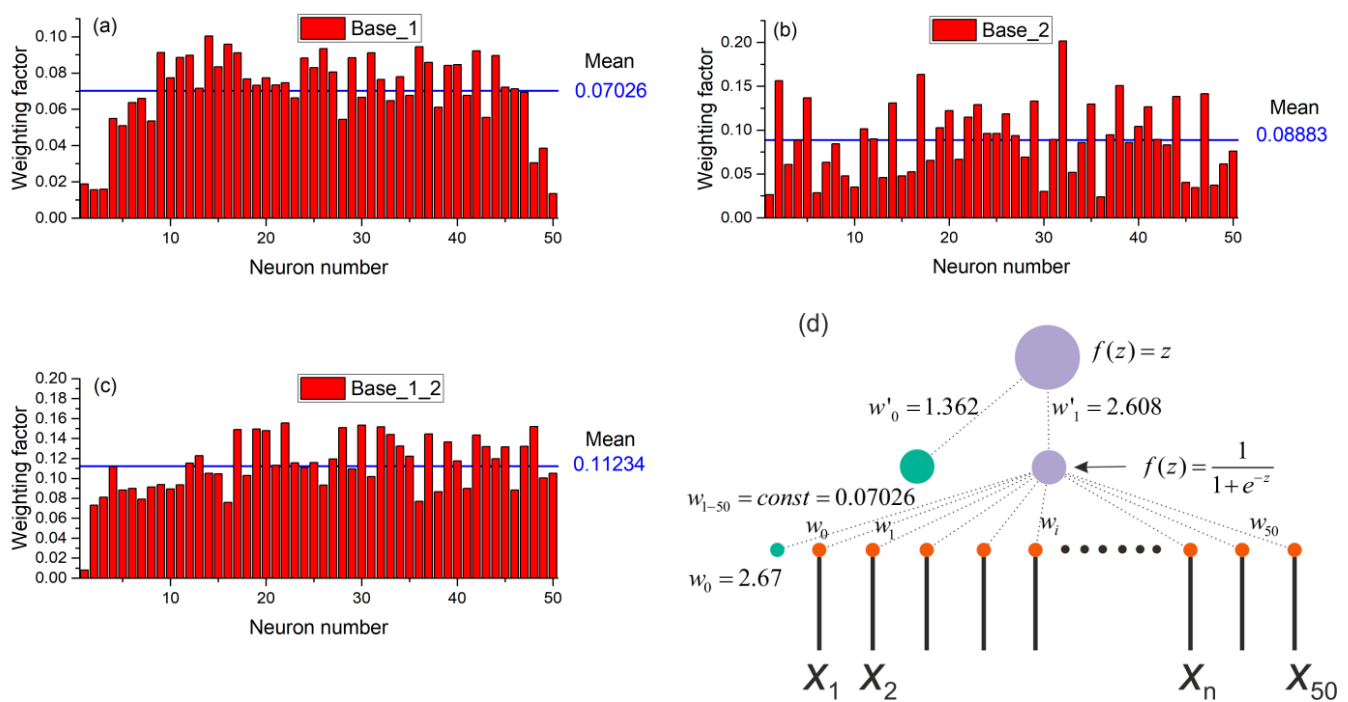

**Figure 12.** Distributions of weights for the perceptron model based on one neuron in the hidden layer for (**a**) Base_1, (**b**) Base_2 and (**c**) Base_1_2. (**d**) Perceptron model with equal weights from input to hidden neuron. The figure shows all the weights of the simplified model of the chaos sensor.

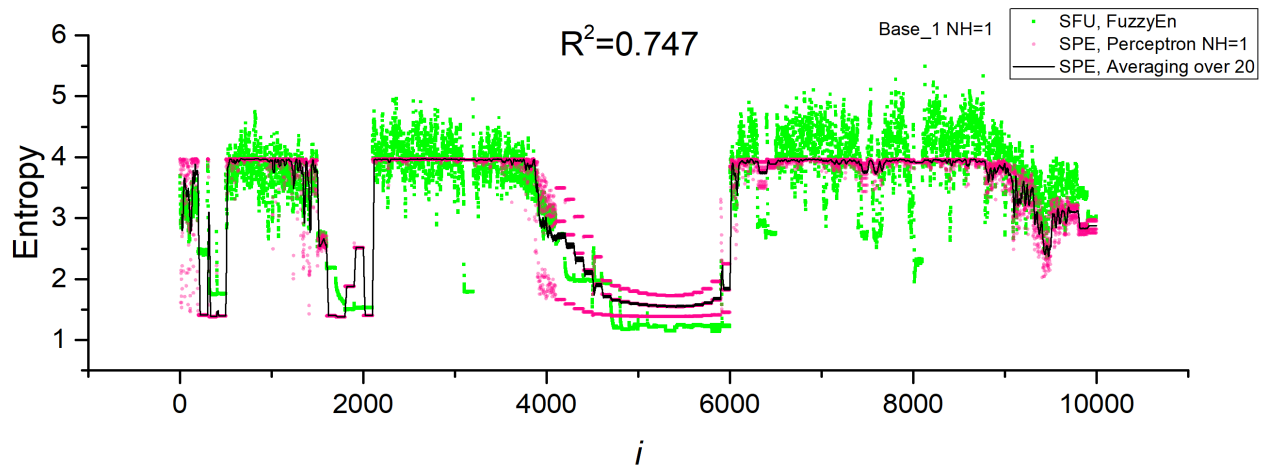

**Figure 13.** Dependences of entropy on the element number in Base_1 for SFU and SPE models, as well as the average value over 20 measurements for SPE (Averaging over 20). The results are shown for a simplified model with one *NH* = 1 neuron and equal weights. Calculations were made on Base_1.

The SPE model with one neuron and weights fitted during training (Figure 9a) is a much better approximation than the simplified model (Figure 13). The SPE model with a large number of neurons in the hidden layer approximates the entropy even better for various combinations of datasets (Table 3). Figure 14 shows a perceptron model with 50 neurons in the hidden layer. The weights from the hidden layer to the output neuron are presented as a histogram in Figure 14a; the distribution of weights from the input layer to the hidden layer is shown in Figure 14b. The general model of the perceptron is shown in Figure 14c. The functioning principle of the sensor on the distribution of weights is difficult to explain and additional research is needed in this direction. Nevertheless, it can be assumed that the input time series is convolved with certain weight patterns, which are also time series, possibly the most frequently occurring in the database, and then the result of the convolution is written to the neurons of the hidden layer. The weights have both positive and negative values, and it is similar to the time series after the normalization step. The value vector of the hidden layer neuron is then convolved with the output layer weights, which also have positive and negative values. In the case of a regular time series input, the result of the convolution has a lower value; in the case of a chaotic series, the result has a higher value.

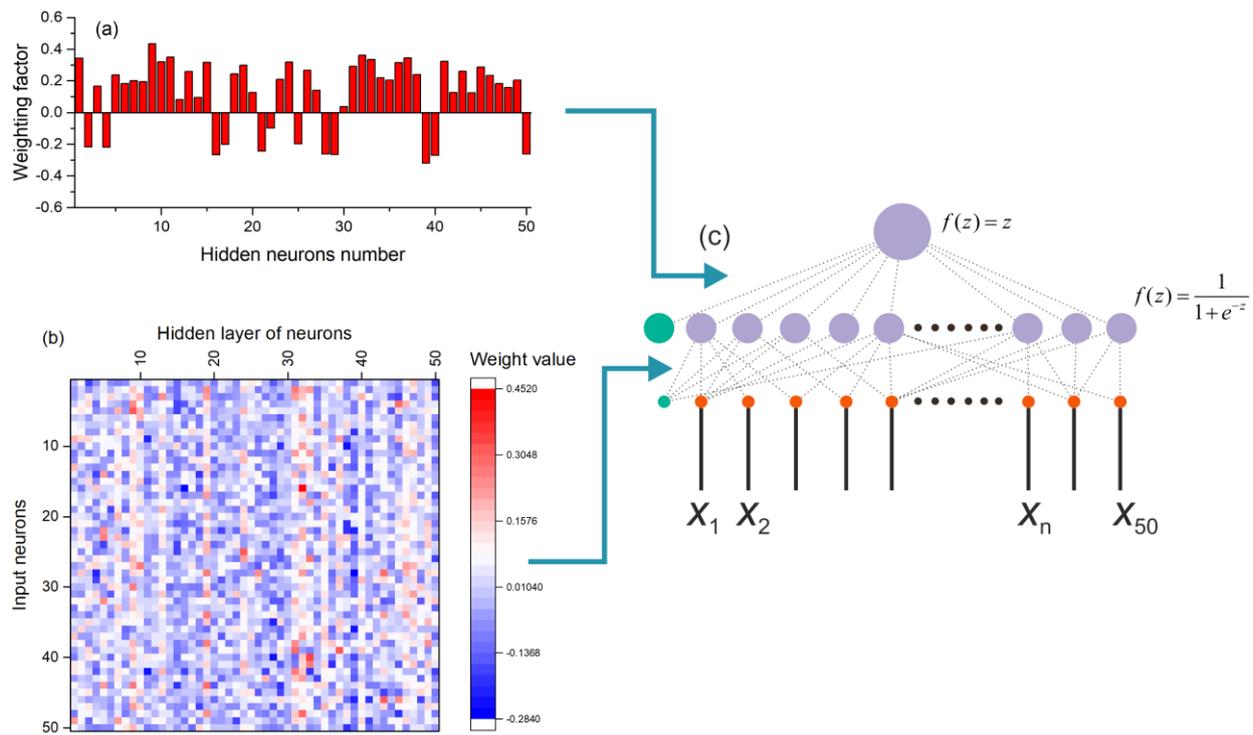

**Figure 14.** Perceptron model with 50 neurons in the hidden layer. The weights from the hidden layer to the output neuron are presented as a histogram (**a**), the distribution of weights from the input layer to the hidden layer is shown in figure (**b**). The general model of the perceptron is shown in Figure (**c**).

*3.4. The Results of Training the Perceptron Sensor of Chaos on Experimental Data*

Section 2.5 described the training dataset (Base_1_exp) and testing dataset (Base_2_exp) obtained from the experimental data of action potentials recorded from the L5 dorsal rootlet of rat [41,76].

Table 5 presents the estimation results of the determination coefficient $R^2$, RMSE and MAPE for perceptron models with the number of neurons in the hidden layer, *NH* = 1, 50 and 150. Preliminary normalization was performed (*Mean* = 0.0013055). The number of neurons in the hidden layer has little effect on the approximation accuracy metrics for the training set Base_1_exp ($R^2 \sim 0.78$). Results for the testing set were higher ($R^2 \sim 0.85$). Figure 15 reflects dependences of entropy on the element number in Base_2_exp for SFU and SPE models. SPE model effectively approximates the entropy of spike train for rest and stimulation modes of the animal. The complete MLP models with weight distribution are presented in Supplementary Materials (Model_2).

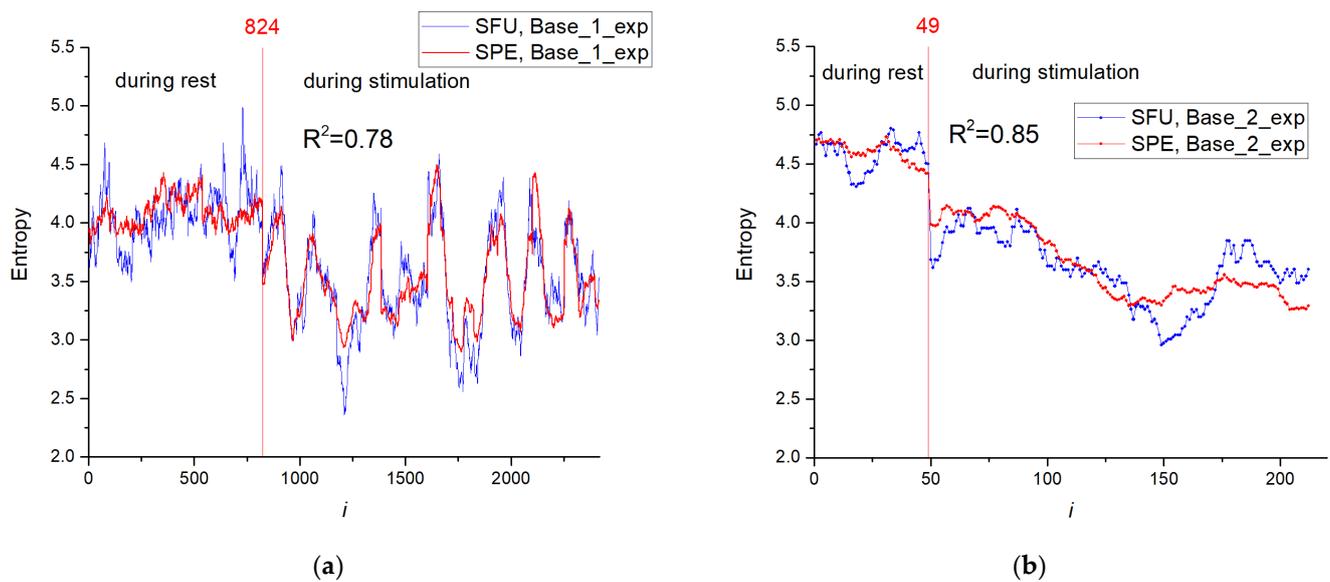

**Figure 15.** Dependences of entropy on the element number in Base_1_exp (**a**) and Base_2_exp (**b**) for SFU and SPE models. The results are shown for a model with *NH* = 50. The efficiency of entropy approximation of spike train by the chaos sensor is presented for two modes corresponding to rest and stimulation of the animal. Spike train of action potentials were recorded from the L5 dorsal rootlet of rat.

**Table 5.** Approximation accuracy estimated by metrics $R^2$, RMSE and MAPE with normalization stage for datasets Base_1_exp and Base_2_exp.

| Number of Neurons in the Hidden Layer, *NH* | Cross-Validation Accuracy for Base_1_exp | | | Approximation Accuracy When Training on Base_1_exp and Testing on Base_2_exp | | |
|---|---|---|---|---|---|---|
| | $R^2$ | RMSE | MAPE, % | $R^2$ | RMSE | MAPE, % |
| 1 | 0.781 | 0.219 | 4.9 | 0.849 | 0.189 | 4.3 |
| 50 | 0.781 | 0.219 | 4.9 | 0.85 | 0.188 | 4.3 |
| 150 | 0.778 | 0.221 | 4.9 | 0.852 | 0.187 | 4.2 |

## 4. Discussion

In biology and medicine, as in other fields of science, there are systems that combine the properties of regular and irregular dynamics and exhibit the properties of deterministic chaos that arises and disappears when the parameters vary. Chaos appears in a variety of studies of wildlife from biochemistry and genetics to population biology and ecology, where chaos can serve as an indicator of instability, development, imminent death, and process characteristic. A question may arise: *Can a simulated model of a biosystem evaluate the randomness of a spike train?* In the course of this study, we demonstrated that it is quite realistic to create a bio-inspired model of a chaos sensor based on a multilayer perceptron. One of the popular spike models used to describe real biological neurons, the Hindmarsh–Rose system, was considered. It allows modeling spike activity in a wide range of spike dynamics with the presence of chaotic and regular modes, with many patterns controlled by the system parameters *r* and $I_{ex}$. Examples of spike waveforms and bifurcation diagrams were demonstrated in Figures 3 and 4. As we tested the perceptron-based sensor within the Hindmarsh–Rose model, we found that the generalizability to other models of spike activity or synthetic time series requires additional research. Nevertheless, in a particular biological system, most likely, the spike activity changes within a certain model system, and it is possible to train the perceptron to distinguish between chaos and order in this particular biosystem (living organism). The question of the actual presence of an ensemble of neurons with a similar function as a chaos sensor may be the subject of future research by neurophysiologists. The paper proposes a bio-inspired tool for assessing chaos, but the question whether the similar mechanism is used in nature remains open. The authors advocate that as a result of evolution, complex combinations of neurons could have arisen that work as a sensor of chaos. Moreover, in an elementary model with one neuron and the same weights (Figure 12), the sensor is the simplest integrator, which can be implemented on other elements, for example, on an integrating RC circuit. The development of these ideas may be a topic for further research.

The idea of a chaos sensor is based on training the perceptron to determine the value of the entropy of the time series entering its input. The time series is a set of distances between spikes with a length of 10 to 100 elements. This approach is model, since it is obvious that nature operates with spikes, and not with distances between them, and in a real biosystem, a time series is not allocated and is not fed to the perceptron as in a computer. Nevertheless, the perceptron model is used widely in neuroscience, and it can qualitatively describe complex processes. In the future, it is possible to work with real spike systems, for example, of a reservoir type, and model the entropy approximation; such studies will be a promising direction for the development of this work.

In this paper, we presented the SFU and SPE models of the chaos sensor. The first model estimates the irregularity of the time series through the calculation of FuzzyEn, and the second model approximates the degree of chaos with a multilayer perceptron. The main goal was to create an SPE model that matches the SFU model as best as possible in terms of the $R^2$ metric, after training the perceptron. The main parameters of the chaos sensor are introduced, such as the sensitivity *EnSens* and the measurement error *EnErr*. FuzzyEn was chosen as a measure of entropy. The selection of its parameters helped to achieve a higher sensitivity of the sensor on a short time series (Figure 8) and a low measurement error. Thus, for the length of the series *NL* = 50, the entropy measurement error by the perceptron was at the level of *EnErr* ~ 11% (Table 4, Sensor on Perceptron). The error can be reduced by averaging the result, for example, averaging over 20 results reduces the measurement error to *EnErr* ~ 4%, almost three times (Table 4). Averaging over the result is equivalent to increasing the length of the time series. And as Figure 8c,d demonstrates, an increase in the length of the series leads to an increase in *EnSens* and a decrease in *EnR*. From a practical point of view, it is more difficult to work with long series, including the perceptron, and averaging can be easily implemented both in model and in real biosystems. Thus, averaging makes it possible to work with even shorter series *NL* = 10, 20. The study of the characteristics of the chaos sensor and the approximation of the perceptron to ultra-short series can be a subject of further research.

We used three variants of MLP with the number of neurons in the hidden layer: *NH* = 1, 50 and 150. MLP with *NH* = 150 has excessive complexity. Although it shows the highest results in cross-validation within the same dataset (see columns Cross-validation accuracy in Tables 2 and 3), this model loses its universality due to retraining during training and testing on different sets and provides lower $R^2$ than MLP at *NH* = 50 (Table 3). The model becomes too specifically tuned to the training data and approximates the test data worse. MLP with *NH* = 50 neurons is the most efficient model compared to other options. Therefore, the study of three MLP variants with 1, 50, and 150 neurons in the hidden layer evaluates the basic properties of the chaos sensor and confirms its effectiveness. The sigmoid activation function used in this study showed its effectiveness for the investigated MLP model. Although this function is common [80], the search for optimal fine-tuning of the model, including the type of activation function, the number of neurons and layers, can be a subject of future study.

As the results of Tables 2 and 3 indicate, the correct normalization of the time series can significantly increase the accuracy of the approximation of the perceptron models. In general, the approach without normalization can be applied in practice if the training base contains all possible combinations of time series, as in the case of Base_1_2. The method of training on one dataset and testing on another dataset showed low $R^2$ values (Table 2, last 2 columns). This indicates that the model is over trained on a separate dataset and is not suitable for working with another dataset. It shows that the two datasets Base_1 and Base_2 have a different set of time series, which is also noticeable in the buffer diagrams (Figure 4) and Table 1, where the *Min_X* and *Max_X* of the two sets differ significantly. This difference was created by us intentionally in order to test the universality of the created perceptron models as a chaos sensor.

The normalization method implies subtraction from the values of the elements of the series the value of the average value *Mean* (Table 1). The results are presented in Table 3 and are significantly better than in the method without normalization. The results suggest that even a single neuron in the hidden layer approximates the entropy well, although the accuracy of the model decreases with an increase in the variety of time series when datasets are combined. This is noticeable when training on one dataset and testing on another set (Table 3, last 2 columns). However, $R^2$ reaches values of $R^2 \sim 0.326$ when training on Base_2 and testing on Base_1, and this result indicates a partial approximation. The reason for the unsatisfactory approximation, in our opinion, is the fact that the principle of operation of a perceptron with one neuron in the hidden layer is related to the average value of the time series (Section 3.3), however, many variants of the time series do not follow this pattern.

The model with normalization and a large number of neurons in the hidden layer $NH$ = 50, 150 approximates the FuzzyEn values with an acceptable accuracy of $R^2 \sim 0.9$ (Figures 9b and 10a). The functioning principle of the sensor, according to the distribution of weights (Figure 14), is difficult to explain and additional research is needed in this direction. It can be assumed that the input time series is rolled up with certain weight patterns representing the most frequently occurring time series in the database. Then, the resulting vector of hidden layer neuron values is convolved with the weights of the output layer, which have positive and negative values. In the case of a regular time series input, the result of the final convolution has a lower value; in the case of a chaotic series, result has a higher value.

The normalization method consists of subtracting the global average from all elements of the time series, and it is equivalent to simple filtering of the constant component. The effectiveness of normalization is determined experimentally and depends on the datasets. For non-stationary time series, when the mean changes over a wide range, more advance methods of preprocessing the series are needed, including the defining of trends from series. Determining the randomness of non-stationary time series may be a topic for future research.

The efficiency of entropy approximation of spike train by the chaos sensor for two modes corresponding to rest and stimulation of the animal is shown in Section 3.4. The spike train of action potentials was recorded from the L5 dorsal rootlet of rat. The use of even one neuron in the hidden layer effectively determines the entropy of impulses with an accuracy of $R^2 \sim 0.78$ for the Base_1_exp and $R^2 \sim 0.85$ for the Base_2_exp (Table 5 and Figure 15). The rest mode of the animal has a higher entropy of impulse action potentials than the stimulation mode. A detailed study and comparison of the obtained results with the results of the original work [41,76], and results of significance from the neurobiology point of view may be a topic for future research with the assistance of specialists in the field.

## 5. Conclusions

The biosimilar model of the chaos sensor based on a multilayer perceptron was proposed and investigated. We tested the perceptron-based sensor within the Hindmarsh–Rose model, and additional studies are required to apply the concept to other models of spike activity or synthetic time series. Overall, we have provided a positive answer to the question: *Can a simulated model of a biosystem evaluate the randomness of a spike train?*

A perceptron model with one neuron in the hidden layer is proposed. The model allows obtaining a high degree of similarity between SFU and SPE models for short series of *NL* = 50 in length. A perceptron model with 50 neurons in the hidden layer is proposed, and it allows obtaining a high degree of similarity between SFU and SPE models with an accuracy of $R^2 \sim 0.9$. An example of using a chaos sensor on spike packets of action potentials recordings from the L5 dorsal rootlet of rat is presented. Potentially, a chaos sensor for short time series makes it possible to dynamically track the chaotic behavior of neuron impulses and transmit this information to other parts of the biosystem. The perceptron model is simple and has a high speed of chaos estimation. In our opinion, the speed of determining the source of chaos in nature can be decisive, for example, when hunting predators, and can be a more important characteristic of the sensor than its accuracy. We hope this study will be useful for specialists in the field of computational neuroscience, and our team is looking forward to interacting with interested parties in joint projects.

**Supplementary Materials:** The following supporting information can be downloaded at: www.mdpi.com/xxx/s1. (Database: Base_1, Base_2, Base_1_2, Base_1_exp, Base_2_exp; Models: Model_1, Model_2); The experimental databases Base_1_exp and Base_1_exp were compiled based on open data of action potentials recorded from the L5 dorsal rootlet of rat [41,76] located at https://data.mendeley.com/datasets/ybhwtngzmm/1 (accessed on 25 July 2023).

**Author Contributions:** Conceptualization, A.V.; methodology, A.V., P.B., M.B. and V.P.; software, A.V., P.B., M.B. and V.P; validation, M.B. and V.P.; formal analysis, A.V.; investigation, A.V., M.B. and V.P.; resources, P.B.; data curation, A.V.; writing—original draft preparation, A.V., P.B., M.B. and V.P.; writing—review and editing, A.V., P.B., M.B. and V.P.; visualization, A.V., P.B., M.B. and V.P.; supervision, A.V.; project administration, A.V.; funding acquisition, A.V. All authors have read and agreed to the published version of the manuscript.

**Funding:** This research was supported by the Russian Science Foundation (grant no. 22-11-00055, https://rscf.ru/en/project/22-11-00055/, accessed on 30 March 2023).

**Institutional Review Board Statement:** Not applicable.

**Data Availability Statement:** The data used in this study can be shared with the parties, provided that the article is cited.

**Acknowledgments:** The authors express their gratitude to Andrei Rikkiev for valuable comments made in the course of the article's translation and revision. Special thanks to the editors of the journal and to the anonymous reviewers for their constructive criticism and improvement suggestions.

**Conflicts of Interest:** The authors declare no conflicts of interest.

## Abbreviations

| | |
|---|---|
| ANN | artificial neural network; |
| CNS | central (cortical) nervous system; |
| ECG | electrocardiography; |
| EEG | electroencephalography; |
| *En*$_{av}$ | average value of entropy; |
| *EnErr* | relative entropy measurement error; |
| *EnR* | range of entropy change; |
| *EnSens* | chaos sensor sensitivity; |
| FuzzyEn | fuzzy entropy; |
| HR | Hindmarsh-Rose; |
| *Mean* | average value for all elements of the time series in the dataset; |
| *Mean50_max* | maximum value of the average value of the time series; |
| *Mean50_min* | minimum value of the average value of the time series; |
| *Max_X* | maximum value of all elements in the dataset; |
| *Min_X* | minimum value of all elements in the dataset; |
| MLP | Multilayer perceptrons; |
| ML | machine learning; |
| *NH* | number of neurons in the hidden layer; |
| *NL* | number of elements of the time series; |
| NNetEn | Neural Network Entropy; |

PermEn permutation entropy;
SampEn sample entropy;
SPE Sensor on Perceptron;
SFU Sensor on fuzzy entropy;
*Std_En* entropy mean square deviation;
SvdEn singular value decomposition entropy.

**Appendix A**

**Table A1.** Estimation of approximation accuracy by RMSE metric without the normalization stage.

| Number of Neurons in the Hidden Layer, *NH* | Cross-Validation Accuracy, RMSE | | | Approximation Accuracy When Training on One Set and Testing on Another Set, RMSE | |
|---|---|---|---|---|---|
| | Base_1 | Base_2 | Base_1_2 | Base_1-Training Base_2-Testing | Base_1-Testing Base_2-Training |
| 1 | 1.12 | 1.25 | 1.2 | 1.27 | 1.14 |
| 50 | 0.462 | 0.452 | 0.52 | 1.25 | 1.16 |
| 150 | 0.434 | 0.426 | 0.473 | 1.16 | 1.51 |

**Table A2.** Estimation of approximation accuracy by MAPE metric without the normalization stage.

| Number of Neurons in the Hidden Layer, *NH* | Cross-Validation Accuracy, MAPE (%) | | | Approximation Accuracy When Training on One Set and Testing on Another Set, MAPE (%) | |
|---|---|---|---|---|---|
| | Base_1 | Base_2 | Base_1_2 | Base_1-Training Base_2-Testing | Base_1-Testing Base_2-Training |
| 1 | 43.9 | 48.5 | 46.9 | 52.2 | 43.1 |
| 50 | 11.7 | 11 | 14.8 | 51.3 | 38.1 |
| 150 | 10.8 | 10.9 | 12.4 | 44 | 52.3 |

**Table A3.** Approximation accuracy estimation by metric RMSE using the normalization stage.

| Number of Neurons in The Hidden Layer, *NH* | Cross-Validation Accuracy, RMSE | | | Approximation Accuracy When Training on One Set and Testing on Another Set, RMSE | |
|---|---|---|---|---|---|
| | Base_1 | Base_2 | Base_1_2 | Base_1-Training Base_2-Testing | Base_1-Testing Base_2-Training |
| 1 | 0.486 | 0.788 | 0.85 | 1.18 | 0.914 |
| 50 | 0.351 | 0.335 | 0.4 | 1.38 | 0.654 |
| 150 | 0.331 | 0.321 | 0.388 | 1.6 | 0.732 |

**Table A4.** Approximation accuracy estimation by metric MAPE using the normalization stage.

| Number of Neurons in the Hidden Layer, *NH* | Cross-Validation Accuracy, MAPE (%) | | | Approximation Accuracy When Training on One Set and Testing on Another Set, MAPE (%) | |
|---|---|---|---|---|---|
| | Base_1 | Base_2 | Base_1_2 | Base_1-Training Base_2-Testing | Base_1-Testing Base_2-Training |
| 1 | 13.5 | 22.3 | 26.7 | 42.6 | 28 |
| 50 | 8.4 | 7.1 | 9.9 | 48.6 | 19.5 |
| 150 | 8.1 | 7 | 9.3 | 55.9 | 21 |